\documentclass[letterpaper, 10 pt, conference]{ieeeconf}  

\IEEEoverridecommandlockouts                              

\usepackage{amsmath} 
\usepackage{amsmath,amsfonts}
\usepackage{algorithmic}
\usepackage{algorithm}
\usepackage{array}
\usepackage{textcomp}
\usepackage{stfloats}
\usepackage{url}
\usepackage{verbatim}
\usepackage{graphicx}
\usepackage{cite}
\usepackage{caption}
\usepackage{multirow}
\usepackage{colortbl} 
\usepackage{xcolor}
\usepackage{subcaption}
\usepackage{arydshln}
\usepackage{placeins}
\usepackage{tabularray}
\UseTblrLibrary{booktabs}
\usepackage{soul}  
\usepackage{caption}
\usepackage{hyperref}

\title{\LARGE \bf
Adaptive Companionship for Group-Following Robots: Handling Dynamically Changing Group Formations
}

\author{Cong-Thanh Vu and Yen-Chen Liu
\thanks{This work was supported in part by the National Science and Technology Council (NSTC), Taiwan, under Grant NSTC 114-2628-E-006-010 and NSTC 114-2218-E-006-021.}
\thanks{Cong-Thanh Vu and Yen-Chen Liu are with the Department of Mechanical Engineering, National Cheng Kung University, Tainan 70101, Taiwan. Email: {\href{mailto:vuthanh.cdt@gmail.com}{\texttt{vuthanh.cdt@gmail.com}}, \href{mailto:yliu@mail.ncku.edu.tw}{\texttt{yliu@mail.ncku.edu.tw}}}.}
}

\begin{document}

\maketitle
\thispagestyle{empty}
\pagestyle{empty}

\begin{abstract}
Accompanying a group of humans is an essential aspect of developing human-like social cognition in robots. However, human groups typically do not follow fixed formations, which poses significant challenges for robots in maintaining natural companionship behaviors. In this paper, we propose an adaptive group-accompaniment method for social robots based on Vision-Language Models (VLMs), leveraging their semantic reasoning capabilities to infer companion positions, maintain social distances, and understand group dynamics. The members of the group are first detected, and a perceptual module generates visual representations of the interaction group space as input to the VLM, which is then combined with a Model Predictive Path Integral (MPPI) controller to ensure stability and safety. Experimental evaluations across five scenarios show that the proposed method enables robots to accompany the group effectively, demonstrating a 15\% improvement in success rate and a 25\% reduction in collision rate compared to baseline approaches. Additionally, a user study indicates that the generated companionship behaviors are perceived as natural and socially appropriate.
\end{abstract}

\section{Introduction}
In recent years, the development of social robots has attracted increasing attention, driven by the growing demand for robots capable of seamless interaction with humans in everyday environments \cite{doi:10.1177/02783649241230562}. Among these capabilities, accompanying an individual or a group of pedestrians is considered a key aspect of human-like social cognition, with a wide range of practical applications, from assisting with daily tasks to providing logistical support \cite{islam2019person,ahmed2024human}.

Over the past two decades, a wide range of companionship strategies have been investigated to enhance the social cognition and interaction capabilities of companion robots. Early research primarily focused on accompanying a single individual, which laid the foundation for companionship tasks in social robotics. Various strategies have been proposed, including following from behind \cite{TOAN2023104317, 9690946, 10606944}, walking side by side \cite{10.5898/JHRI.3.2.Morales, 10947582, peng2023mpc, iccre}, and leading in front of the human \cite{9561974, 10869380}. In addition to improving tracking, socially aware navigation frameworks that respect interpersonal distances \cite{10611263, repiso2024} and learning-based methods that adapt to individual walking patterns \cite{11246444,10869380} have been developed to enhance naturalness and human comfort. More recently, human-companion strategies have been extended to quadruped robots, enabling more agile and terrain-adaptive behaviors \cite{11127912}.
\begin{figure}[t]
\centerline{\includegraphics[scale=0.085,page=1]{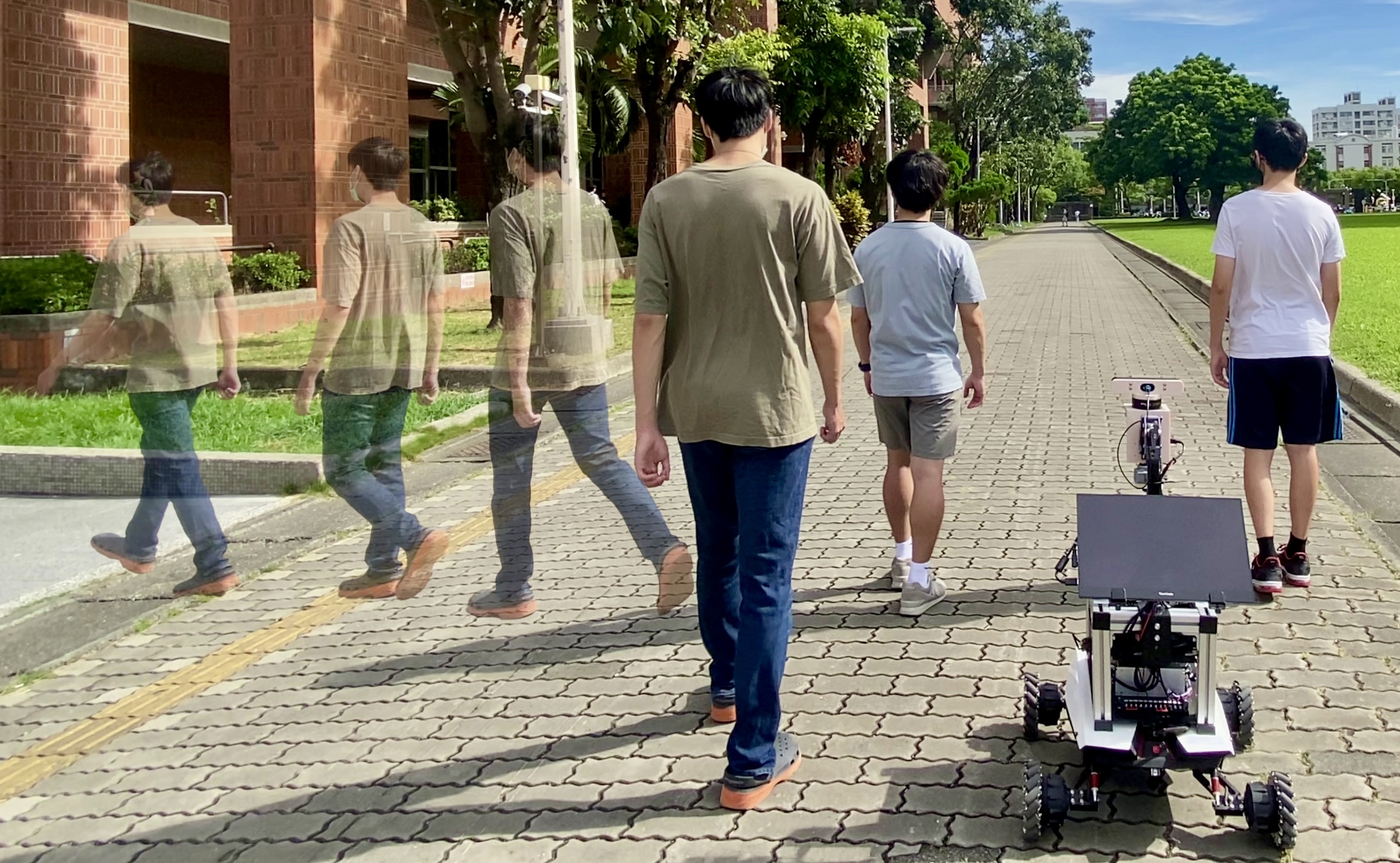}}
\caption{A member leaves the group, changing the group formation.}
\label{fig:example}
\end{figure}

In everyday life, humans often move not only individually but also in groups \cite{10265759}. For robots, accompanying pedestrian groups poses significant challenges due to the dynamic and often unpredictable behaviors of multiple individuals. Groups of two often walk side by side to facilitate interaction, while groups of three commonly adopt a V-shaped formation \cite{PhysRevE.89.012811}. However, group members are not always arranged in such fixed formations, and their interactions generate complex social dynamics that the robot must recognize and adapt to, including constraints on available walking space \cite{zanlungo2019intrinsic}. Moreover, groups in social spaces are inherently dynamic, as individuals may join or leave at any time, as shown in Fig.~\ref{fig:example}, which further increases the complexity of the task. Therefore, effective group accompaniment requires adaptive strategies that go beyond single-person tracking, integrating social awareness, spatial reasoning, and dynamic decision-making to ensure natural and safe interactions.

Although research on human-following robots has been conducted for quite some time, most previous studies have primarily focused on accompanying a single individual. Research on human groups has typically emphasized group detection to facilitate navigation and collision avoidance \cite{10918817}, while other studies have investigated the synchronization of group movements to generate robot behaviors for coordinated tasks, such as integrating robots into dance performances \cite{7494678}.

Only a few studies have addressed the challenge of enabling robots to accompany pedestrian groups. Repiso et al. \cite{8968601} proposed an adaptive group-following approach that uses a social force model to predict human behavior, combined with movement patterns such as V-shaped or side-by-side formations. This approach was later extended \cite{8976304,repiso2024adaptive} to improve robot behavior and performance, allowing the robot to prioritize side-by-side accompanying behavior. However, the method still requires the user’s goal to be provided, assumes fixed group membership, and only considers groups of two individuals. In addition, the reliance on fixed formations, such as V-shaped or side-by-side, limits applicability because real pedestrian groups do not consistently maintain these structures. Leader-selection strategies within a group were also explored by Liao et al. \cite{11123734}. However, situations in which the formation is disrupted, such as when a person leaves or a new individual joins, have not yet been examined.

Recent breakthroughs in Large Language Models (LLMs) and Vision-Language Models (VLMs) have demonstrated remarkable abilities in deep contextual understanding, commonsense reasoning, and multi-step inference, reflecting characteristics of human thought processes. These capabilities have been applied across diverse robotics domains. Their reasoning abilities have also enabled applications in social robotics, such as navigation guided by human reasoning \cite{10777573,11128004} and group detection for movement decision-making \cite{11106758}. Furthermore, they have been applied to complex agricultural environments \cite{VU2026112092}. However, the use of language models in companion robotics, particularly for the group-following task, has received limited attention to date.
\begin{figure*}[t]
\centering
\includegraphics[scale=0.29,page=1]{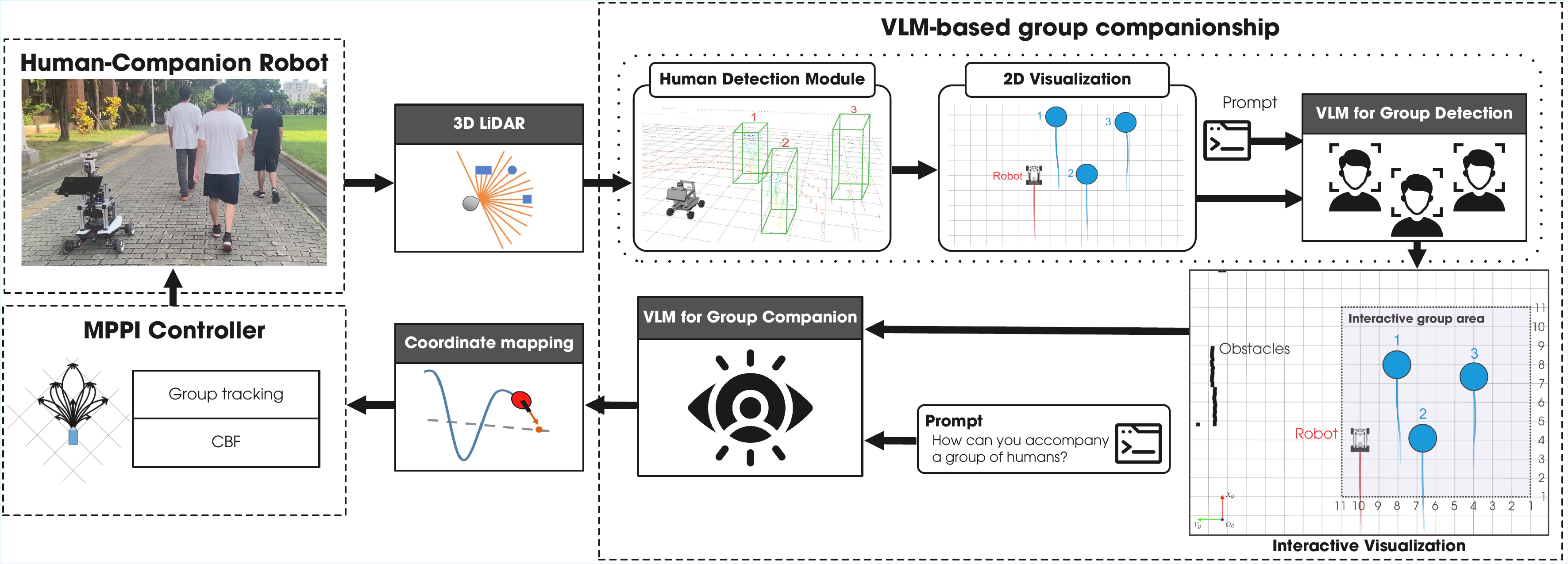}
\caption{The overall structure of the system architecture. VLM uses a visual representation of detected humans and the robot to infer the companion group. Based on the identified group IDs, a grid-based interaction space is constructed, enabling the VLM to reason about adaptive companion positions. These positions are then provided as input to the MPPI controller.}
\label{fig:diagram}
\end{figure*}

In this paper, we introduce a novel group-companion robot approach that adaptively adjusts its position without relying on pre-defined formations. The method can handle disruptions in group membership, such as when someone leaves or joins, which has not been addressed before. The main contributions are:
\begin{enumerate}
\item We present an integration of VLM-based reasoning into the companion task for human-group following. By leveraging human-like inference capabilities, the robot can select appropriate positions within the group and recognize or anticipate membership changes, thereby improving the robustness of group-following behaviors.
\item We propose a group detection module based on VLM, which allows the robot to recognize when a member leaves or a new member joins, based on proxemic distances and path similarity. In addition, the group interaction space is introduced, enabling the VLM to make accompaniment decisions within this space rather than issuing inappropriate actions.
\item We validate the proposed approach through five scenarios, comparing it with the methods in \cite{iccre,8976304,11123734}, followed by a user study and an ablation study. Experimental results demonstrate a 15\% improvement in success rate and a 25\% reduction in collision rate. Furthermore, the user study confirms an enhanced user experience when interacting with the robot.
\end{enumerate}
\section{Problem Formulation}\label{sec:system}
Consider the robot’s state $\mathbf{q}^r = [x^r, y^r, \theta^r]^\text{T} \in \mathbb{R}^3$ in the global frame $\mathcal{G}$. The robot’s task is to determine the optimal accompaniment position $\mathbf{q}^d = [x^d, y^d]^\text{T}\in \mathbb{R}^2$  for following a pedestrian group $G$ with $k$ members, whose positions are $\mathbf{q}^h_{i} = [x^h_{i}, y^h_{i}, \theta^h_{i}]^\text{T} \in \mathbb{R}^3$ for $i = 1, \ldots, k$ in the frame $\mathcal{G}$. The robot perceives its environment through point cloud data $P$ obtained from a 3D LiDAR sensor. The accompaniment position $\mathbf{q}^d$ must ensure safety by avoiding collisions with both static obstacles and group members, while adapting to potential changes in the group inferred from observations. The robot’s current position is then adjusted via an accompaniment controller to reach $\mathbf{q}^d$. We use a VLM to predict the group members’ behavior, allowing the robot to understand when members leave or new members join the group. Based on the VLM’s predictions, the robot determines its accompaniment position using environmental information, reasoning in a manner analogous to how humans select positions to move alongside their companions.
\section{Method}\label{sec:method}
This section presents the proposed approach, which leverages a VLM to detect group members and determine the robot’s companion position, followed by an accompaniment controller based on MPPI. The overall structure of the method is illustrated in Fig.~\ref{fig:diagram}.
\subsection{Companion Group Identification}
The first task for a robot to accompany a group of people is to identify the individuals within the group $G$. Unlike previous studies on group companionship \cite{8968601,8976304,repiso2024adaptive}, which assume that group members are known in advance and remain unchanged, detecting changes, such as someone joining or leaving the group, remains a significant challenge. To address this issue, we propose a VLM-based method for group member identification, leveraging its capability to process multimodal inputs and support high-level semantic understanding and reasoning.

\begin{figure}[h]
\centerline{\includegraphics[scale=0.23,page=1]{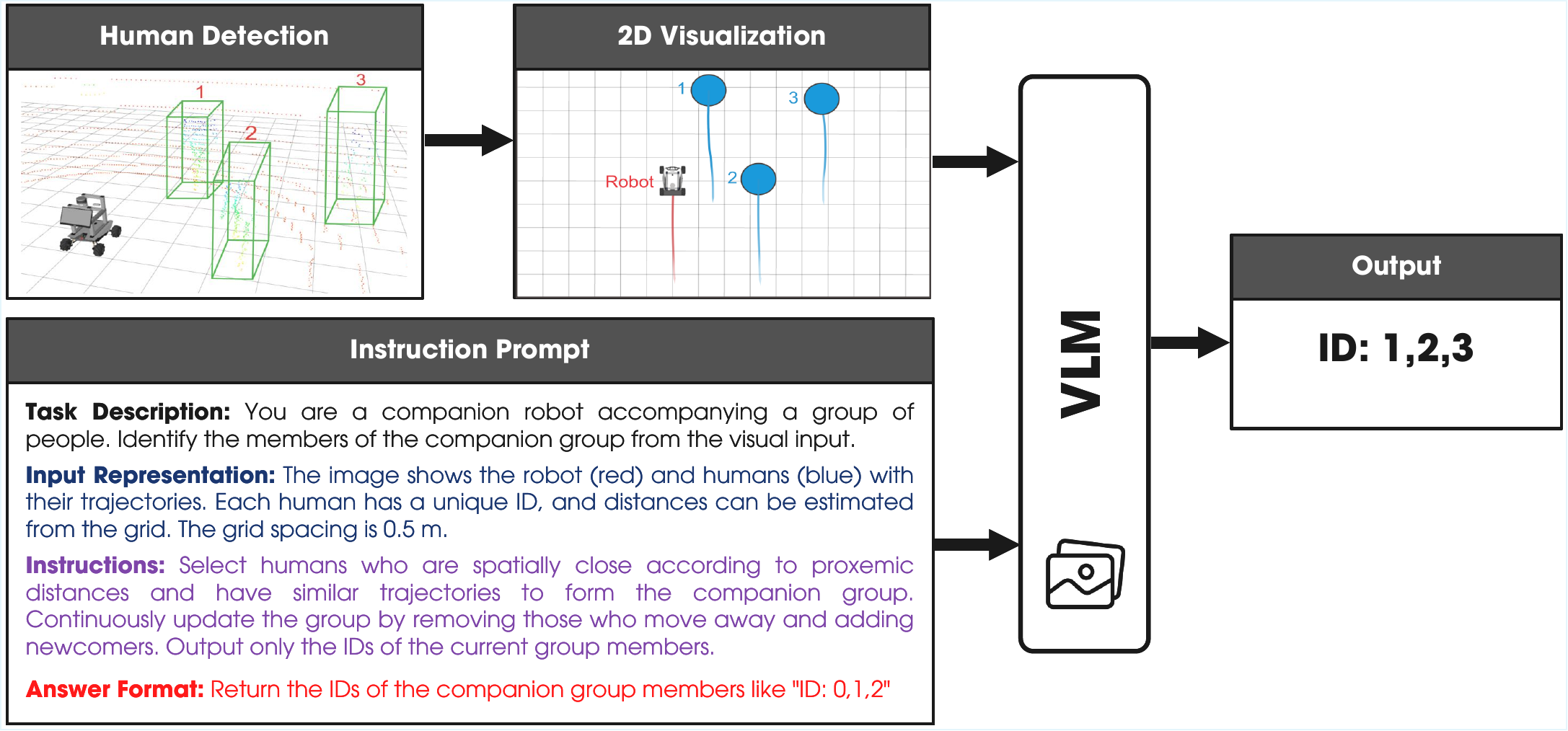}}
\caption{VLM identifies members of the companion group from an image representing human and robot trajectories.}
\label{fig:group_detect}
\end{figure}

Individuals are detected using 3D LiDAR point cloud data processed with PointPillars \cite{lang2019pointpillars}. Each detected object is assigned a unique ID and projected into a 2D space that encodes the robot’s position, individual trajectories, and corresponding IDs. Although VLMs have been applied to social group recognition \cite{11106758}, prior work is constrained by the narrow field of view of camera-based inputs, making it unsuitable for continuous group accompaniment. In contrast, our method incorporates trajectory similarity and proxemic distances within a 2D visual representation, which is provided to the VLM. This enables the model to infer group membership based on trajectories and spatial relations with the robot, rather than relying on camera images limited by field of view. Furthermore, arranging the representation on a grid allows the model to explicitly reason about the relative positions of individuals. Formally, group detection $G$ by the VLM is expressed as:
\begin{align}
G &= \text{VLM}_{\text{group}}(I_{\text{group}}, \mathcal{P}_\text{group}),
\end{align}
where $I_{\text{group}}$ is the visual image representation reconstructed from the positions of the robot and humans, annotated with their IDs and trajectories, and $\mathcal{P}_\text{group}$ is an instruction prompt that defines the context and guides the model to perform the task and generate the desired output. The designed prompt is illustrated in Fig.~\ref{fig:group_detect}.
\subsection{VLM-Based Group Companionship}
Although VLMs exhibit human-like reasoning capabilities, they remain limited when relying solely on visual input, particularly for tasks such as inferring spatial positions directly from RGB images or performing complex mathematical computations. Thus, we limit direct computations and transformations for the VLM by introducing a grid-based group interaction space, which defines the positions where the robot can operate and accompany the group during movement. This framework allows the VLM to reason over predefined candidate positions rather than processing raw visual information or performing mathematical conversions. The group space is represented as:
\begin{align}
\mathcal{Q} &= \left\{ \mathbf{q}^d_{n,m} \;\middle|\; n = 1, \dots, N,\; m = 1, \dots, M \right\},
\end{align}
where each grid position is given by
\begin{align}
\mathbf{q}^d_{n,m} &=
\begin{bmatrix}
x_{\min}(G) + (n-1)\Delta_{\text{grid}} - x_{\text{offset}} \\
y_{\min}(G) + (m-1)\Delta_{\text{grid}} - y_{\text{offset}}
\end{bmatrix}. \label{eq:convert}
\end{align}

The number of columns $N$ and rows $M$ is determined from the group’s spatial extent and the grid spacing $\Delta_{\text{grid}}$ as:
\begin{align}
N &=  \frac{x_{\max}(G) - x_{\min}(G) + 2x_{\text{offset}}}{\Delta_{\text{grid}}} + 1, \\
M &= \frac{y_{\max}(G) - y_{\min}(G) + 2y_{\text{offset}}}{\Delta_{\text{grid}}} + 1.
\end{align}
Here, $x_{\min}(G)$, $x_{\max}(G)$, $y_{\min}(G)$, and $y_{\max}(G)$ denote the minimum and maximum $x$ and $y$ coordinates of the group, while $x_{\text{offset}}$ and $y_{\text{offset}}$ adjust the overall placement of the grid.

The group interaction space $\mathcal{Q}$ is represented as a visual input to the VLM. This input encodes information about the group, including the positions, identifiers, and trajectories of individual members, as well as the position and trajectory of the robot. Two-dimensional environmental obstacles $O$ are also incorporated by projecting a layer of point cloud data $P$, providing spatial context and enhancing the information available for the robot’s decision-making. To avoid random position selection by the VLM, the grid cells are indexed in a structured manner with $N$ columns and $M$ rows. Accordingly, the selection of a position within the interaction space $\mathcal{Q}$ can be expressed as:
\begin{align}
(n,m) = \text{VLM}(I, O, N, M, \mathcal{P}),
\end{align}
where $I$ represents the visual input, and $\mathcal{P}$ denotes the prompt used to guide the model’s reasoning. This indexing scheme enables the VLM to systematically reference and select candidate positions, which are subsequently converted into spatial coordinates using \eqref{eq:convert}.

\begin{figure}[h]
\centerline{\includegraphics[scale=0.31,page=1]{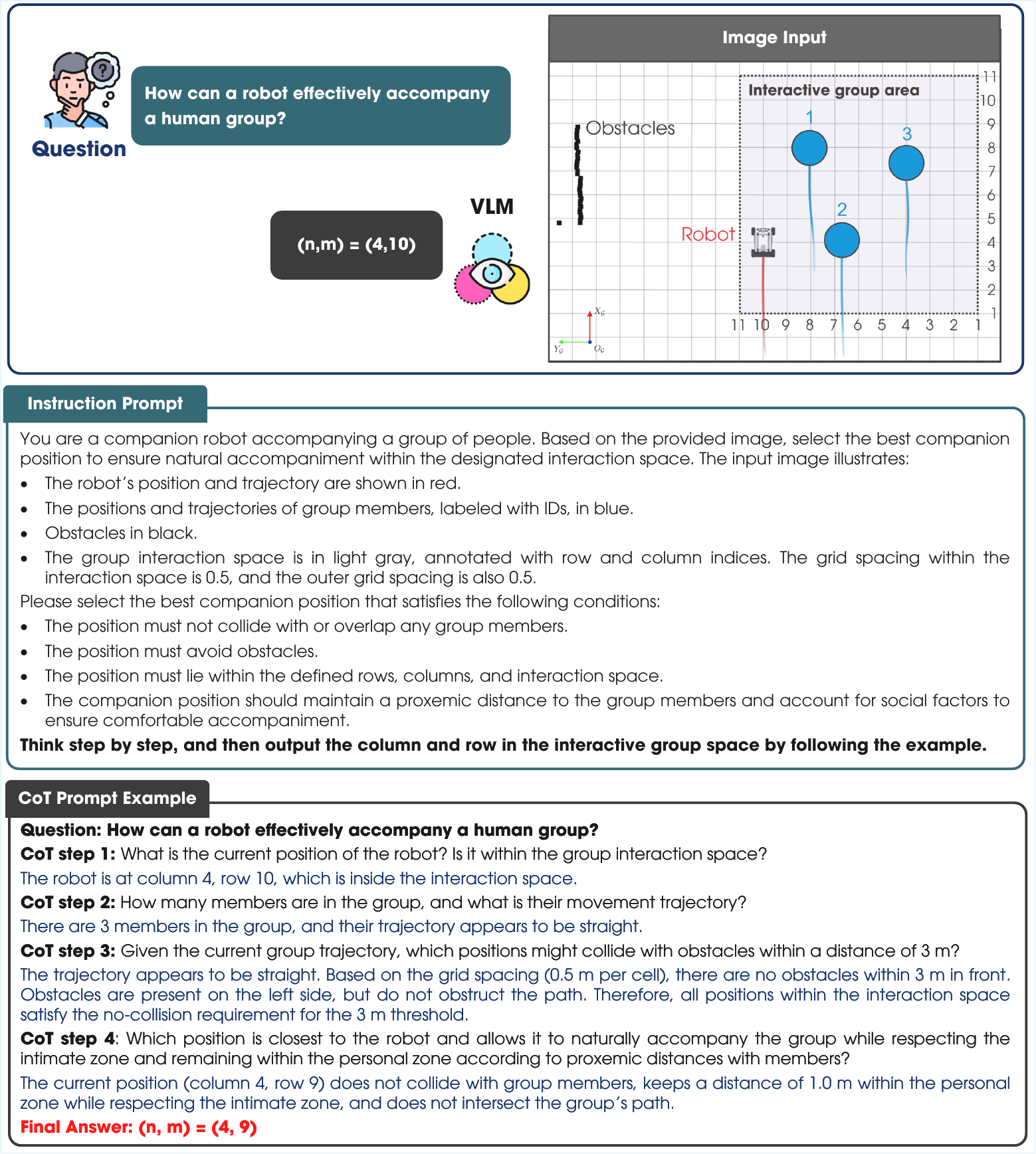}}
\caption{CoT prompting with a one-shot example enables the VLM to perform step-by-step reasoning for the task of accompanying human groups, thereby providing row and column positions within the interaction space.}
\label{fig:companion_prompt}
\end{figure}

To enable the VLM to reason and identify tasks effectively, we employ prompt engineering based on Chain-of-Thought (CoT) prompting combined with a one-shot example. This approach guides the VLM to perform step-by-step reasoning according to predefined structures, similar to human reasoning, while avoiding irrelevant information. This technique allows the robot to systematically identify potential positions within the designed interaction space, such as selecting locations that adhere to social norms, avoid obstructing group members, maintain stable companionship, or adjust when the human formation changes. Moreover, the robot can incorporate information about environmental obstacles using the grid distances to estimate collision risks and eliminate unsafe positions. The CoT prompt for the companion task with the human group is illustrated in Fig.~\ref{fig:companion_prompt}.
\subsection{Human-Companion Controller}
The MPPI (Model Predictive Path Integral) controller, combined with a Control Barrier Function (CBF), is used as the group-tracking controller with the desired tracking position $\mathbf{q}^d$, adapted from \cite{yin2023shield}. The cost functions for the controller are defined as follows:
\begin{align}
&C_{\text{tracking}} = w_x(x^r - x^d)^2 + w_y(y^r - y^d)^2, \\ 
&C_{\text{obs}} = w_{o} 
\begin{cases} c_{\text{collision}}, & \text{if } l(\mathbf{q}^{r}) \ge \tau \\ 
0, & \text{otherwise} 
\end{cases}, 
\end{align}
where the weighting factors $w_{x}$, $w_{y}$, and $w_{o}$ correspond to tracking along the $x$- and $y$-axes and to obstacle avoidance, respectively. Here, $\tau$ is a threshold used to detect potential collisions based on the local costmap. The function $l(\mathbf{q}^r)$ returns the cost value from the local costmap constructed using LiDAR data.

Although the cost term $C_{\text{obs}}$ helps avoid obstacles, it does not explicitly guarantee the preservation of users’ intimate zones or the safety of group members. Therefore, a CBF is employed, with a barrier function $h(\cdot)$ defined for each group member as:
\begin{align}
h(\mathbf{q}^r)_i & = (x^r - x^h_i)^2 + (y^r - y^h_i)^2 - r^2,
\end{align}
where $r$ denotes the minimum allowable distance defining the boundary of each user’s intimate space.

The MPPI controller is implemented in Python using PyTorch to leverage GPU-accelerated parallel processing. The parameters are set as follows: $x_{\text{offset}} = 1.5$, $y_{\text{offset}} = 1.5$, $\Delta_{\text{grid}} = 0.5$, $w_x = 50$, $w_y = 10$, $w_o = 10000$, $\tau = 50$, and $r = 1.0$.
\section{Results and Discussion}\label{sec:results}
In this section, the proposed method is validated through evaluations of the VLM-based group-following strategy and comparisons with previous studies. Five controlled scenarios were designed, complemented by a user study to assess the system’s effectiveness in realistic settings. An ablation study was further conducted to analyze the contribution of each component within the proposed framework. In addition, real-world experiments with extended group configurations were performed to demonstrate the robustness and scalability of the approach.
\begin{figure}[h]
\centerline{\includegraphics[scale=0.4,page=1]{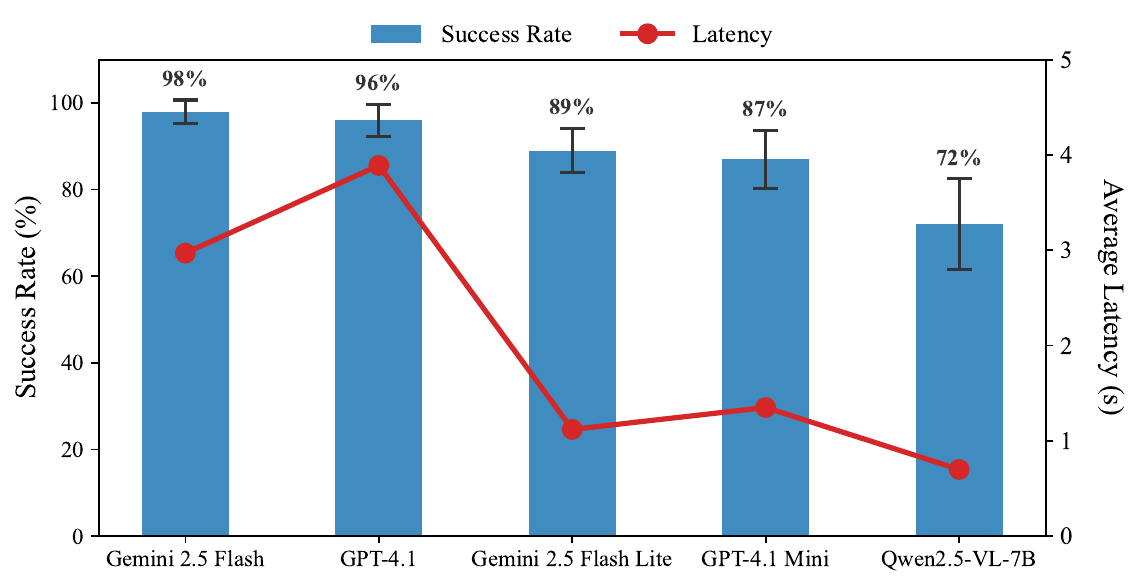}}
\caption{Performance comparison of different VLMs for the group-following strategy.}
\label{fig:vlm_performane}
\end{figure}
\subsection{Evaluation of VLM-Based Group Accompaniment Strategy}
To verify that the VLM-based accompaniment positions are both socially appropriate and physically navigable, we evaluated the positions generated by the VLM across 30 distinct group-accompaniment scenarios. These scenarios included dynamic situations such as new members joining the group or existing members leaving the formation, with group sizes ranging from 5 to 10 individuals. Five different VLM architectures were used to assess the accuracy and overall performance of the proposed method, namely Gemini 2.5 Flash Lite, GPT-4.1 Mini, Gemini 2.5 Flash, GPT-4.1, and Qwen2.5-VL-7B. Each scenario was repeated 10 times for every model.

A model is considered successful if the predicted position satisfies all three evaluation criteria: the proposed coordinates fall within the predefined valid accompaniment region, the selected position avoids physical interference with surrounding individuals or obstacles, and the robot maintains a smooth and socially appropriate formation within the group.
\begin{figure*}[t]
\centering
\includegraphics[scale=0.19,page=1]{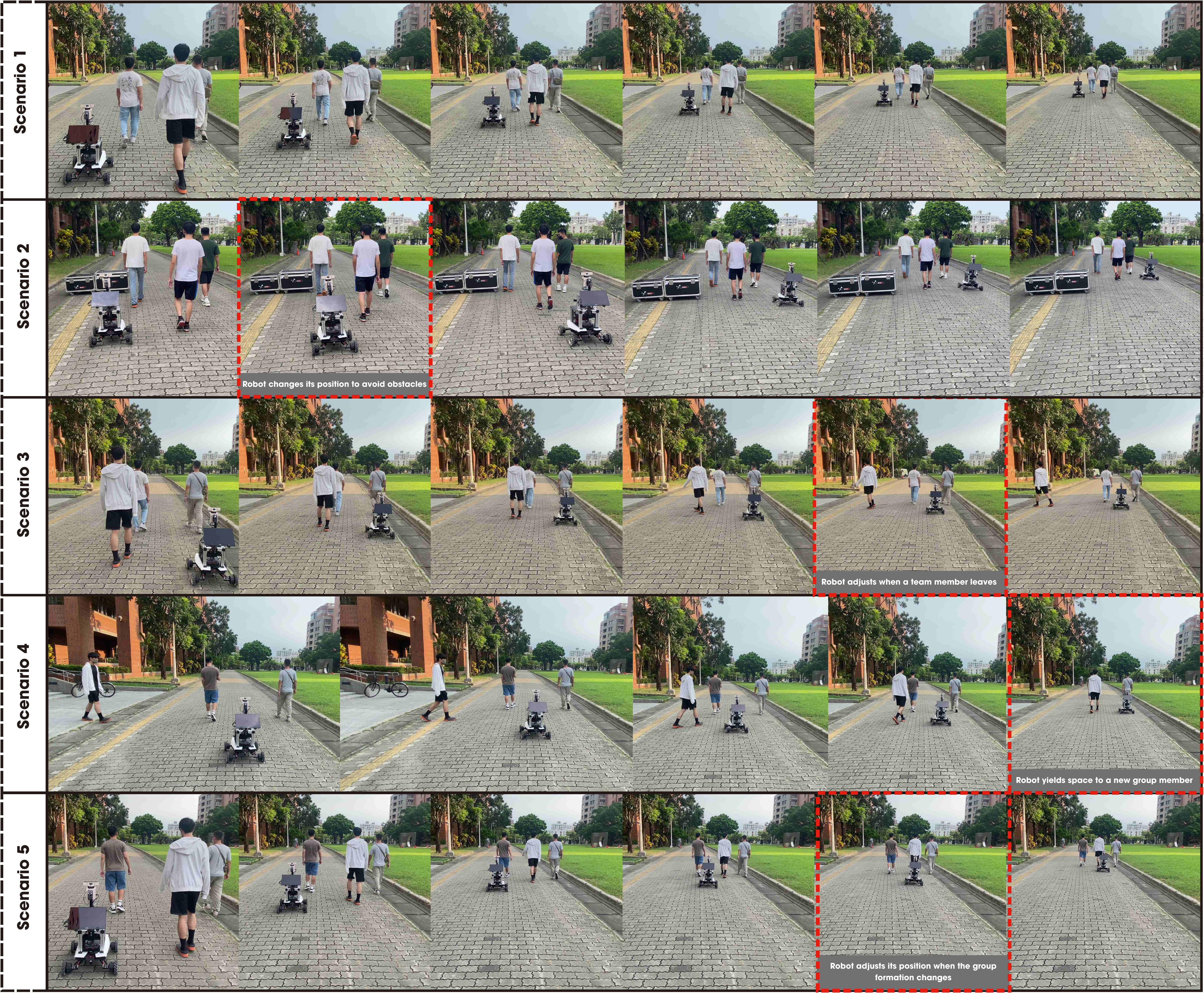}
\caption{Qualitative results illustrating the motion of the proposed method across the five scenarios.}\label{fig:motion}
\end{figure*}

The experimental results, illustrated in Fig.~\ref{fig:vlm_performane}, show that the larger models, Gemini 2.5 Flash and GPT-4.1, achieved the highest success rates, with averages of $98 \pm 1.2\%$ and $96 \pm 1.5\%$, respectively. The lighter variants also performed competitively, with Gemini 2.5 Flash Lite reaching $89 \pm 2.1\%$ and GPT-4.1 Mini achieving $87 \pm 2.7\%$. Regarding computational efficiency, we observed a clear trade-off between model complexity and latency. Gemini 2.5 Flash Lite achieved the lowest average response time at 1.12 s, followed by GPT-4.1 Mini at 1.35 s. In contrast, Gemini 2.5 Flash required 2.97 s on average, and GPT-4.1 exhibited the highest latency at 3.89 s. Notably, the on-board model Qwen2.5-VL-7B provided the fastest inference at approximately 0.7 s, but this came with a substantial reduction in accuracy, achieving only $72 \pm 10.5\%$. These findings suggest that although the larger VLMs offer more robust reasoning capabilities, the Lite and Mini variants provide a desirable balance between accuracy and latency without requiring any fine-tuning. Meanwhile, the on-board Qwen2.5-VL-7B model demonstrates promising responsiveness, but its accuracy indicates that task-specific fine-tuning would be necessary to achieve comparable reliability. Therefore, we select Gemini 2.5 Flash Lite as the VLM for the remaining evaluations, balancing performance and response time.
\subsection{Comparative Evaluation}
\subsubsection{Experiment Setup}
The proposed method was evaluated on an omnidirectional mobile robot platform, the AgileX Robotics Scout Mini, equipped with a Livox MID-360 3D LiDAR sensor. The algorithms were implemented in ROS 2 Humble and executed on an NVIDIA Jetson Orin, leveraging CUDA for accelerated computation. To assess performance, we compared our method against ESFM \cite{8976304}, a side-by-side group-accompaniment approach, as well as the People-as-Planner (PP) method \cite{11123734}, which selects a leader within the group to follow. Additionally, we evaluated it against MPPI \cite{iccre}, an individual-following baseline, to investigate whether behaviors designed for single-person accompaniment can generalize to group settings. All methods were tested across the five scenarios illustrated in Fig.~\ref{fig:motion}:
\begin{itemize}
    \item \textbf{Scenario 1:} Evaluates the baseline tracking performance of the robot when accompanying a group of three people.
    \item \textbf{Scenario 2:} Evaluates performance when obstacles are present while accompanying three people.
    \item \textbf{Scenario 3:} Evaluates the robot’s ability to detect group changes and adapt its accompanying position when one member leaves a group of three.
    \item \textbf{Scenario 4:} Evaluates the robot’s ability to detect group changes and adapt its position when accompanying two people with a new member joining the group.
    \item \textbf{Scenario 5:} Evaluates the robot’s ability to adjust its accompanying position as three group members move and change formation.
\end{itemize}
\begin{table*}[t]
\centering
\caption{Quantitative Results of Comparison}\label{tab:comparison}
\setlength{\tabcolsep}{2pt} 
\small 
\begin{tabular}{l 
    *{5}{>{\centering\arraybackslash}p{0.57cm} 
          >{\centering\arraybackslash}p{0.57cm} 
          >{\centering\arraybackslash}p{1.65cm}} 
} 
\toprule
\multirow{2}{*}{\textbf{Method}} 
& \multicolumn{3}{c}{\textbf{Scenario 1}} 
& \multicolumn{3}{c}{\textbf{Scenario 2}} 
& \multicolumn{3}{c}{\textbf{Scenario 3}} 
& \multicolumn{3}{c}{\textbf{Scenario 4}} 
& \multicolumn{3}{c}{\textbf{Scenario 5}} \\
\cmidrule(lr){2-4} \cmidrule(lr){5-7} \cmidrule(lr){8-10} \cmidrule(lr){11-13} \cmidrule(lr){14-16}
& SR$\uparrow$ & CR$\downarrow$ & CD$\uparrow$
& SR$\uparrow$ & CR$\downarrow$ & CD$\uparrow$
& SR$\uparrow$ & CR$\downarrow$ & CD$\uparrow$
& SR$\uparrow$ & CR$\downarrow$ & CD$\uparrow$
& SR$\uparrow$ & CR$\downarrow$ & CD$\uparrow$ \\
\midrule
MPPI~\cite{iccre} & 100 & 0 & $0.75 \pm 0.07$
& 20 & 80 & $0.51 \pm 0.14$
& 0 & 0 & $2.31 \pm 0.12$
& 50 & 40 & $0.59 \pm 0.16$
& 30 & 70 & $0.41 \pm 0.13$ \\
\addlinespace[2pt]
ESFM~\cite{8976304} & 100 & 0 & $0.81 \pm 0.13$
& 30 & 60 & $0.62 \pm 0.17$
& 10 & 0 & $2.97 \pm 0.27$
& 35 & 60 & $0.52 \pm 0.15$
& 0 & 85 & $0.45 \pm 0.07$ \\
\addlinespace[2pt]
PP~\cite{11123734} & 100 & 0 & $1.34 \pm 0.27$
& 75 & 10 & $1.71 \pm 0.35$
& 55 & 0 & $1.67 \pm 0.23$
& 60 & 35 & $1.03 \pm 0.25$
& 65 & 30 & $1.34 \pm 0.21$ \\
\addlinespace[2pt]
\rowcolor{gray!20} \textbf{Proposed} & \textbf{100} & \textbf{0} & \boldmath{$1.1 \pm 0.04$}
& \textbf{80} & \textbf{10} & \boldmath{$1.05 \pm 0.06$}
& \textbf{90} & \textbf{0} & \boldmath{$1.07 \pm 0.09$}
& \textbf{85} & \textbf{0} & \boldmath{$1.03 \pm 0.07$}
& \textbf{95} & \textbf{5} & \boldmath{$0.87 \pm 0.11$} \\
\bottomrule
\end{tabular}
\end{table*}
Three quantitative metrics were used for evaluation, with results averaged over 20 runs. Success Rate (SR) denotes the percentage of successful accompaniment episodes, where the robot is considered to have failed if it lags more than 2.5 m behind the group for longer than 5 s. Collision Rate (CR) represents the percentage of episodes involving collisions, while Comfortable Distance (CD) measures the average distance maintained between the robot and group members, reflecting its adherence to personal space.
\subsubsection{Comparison Results}
The comparison results, presented in Table~\ref{tab:comparison}, show that the proposed method improved the average success rate by at least 15\% and reduced the collision rate by 25\%. The comfortable distance remained within 0.87–1.1 m, demonstrating greater stability compared with MPPI, ESFM, and PP.

Qualitative outcomes are also illustrated in Fig.~\ref{fig:motion}, which presents the motion sequences of the proposed method across the five scenarios. In Scenario 1, all methods successfully completed the task, as the robot moved straight without changes in group formation or obstacles. In Scenario 2, MPPI failed because it misidentified a group member as an obstacle while switching sides to avoid obstacles, resulting in unstable trajectories and collisions. ESFM maintained its side-by-side configuration and avoided obstacles but ultimately collided due to the narrow passage. Notably, the proposed method and PP completed the scenario with higher success rates and fewer collisions.

In Scenario 3, the proposed method successfully detected a member leaving the group and continued accompanying the remaining two members, while MPPI and PP followed the departing individual. ESFM, unable to detect the departure, relied on the group centroid for tracking, which caused it to lose the group entirely. These results highlight the limitations of applying single-person tracking methods such as MPPI or leader-selection approaches such as PP to group-level tracking, as well as the inability of ESFM to adapt to changes in group composition.

In Scenario 4, the proposed method correctly recognized the arrival of a new member and yielded space accordingly, while MPPI, PP, and ESFM all treated the new member as an obstacle, resulting in collisions. Finally, in Scenario 5, when the group changed formation, the proposed method maintained an appropriate accompanying position without intruding into the group, unlike ESFM, or shifting excessively to the side, like MPPI. Both of these behaviors resulted in collisions with other members.
\begin{table}[t]
\centering
\caption{User Study Questionnaire}\label{tab:survey}
\begin{tabular}{c!{\vrule}p{7.5cm}} 
\toprule
\multicolumn{2}{l}{\textbf{Comfort Scale}} \\
\midrule
1 & I felt comfortable when the robot was following me. \\
2 & I felt safe when the robot was accompanying me. \\
3 & The robot maintained an appropriate distance from me. \\
\midrule
\multicolumn{2}{l}{\textbf{Sociability Scale}} \\
\midrule
4 & The robot’s behavior while following me seemed natural. \\
5 & The robot responded socially appropriately. \\
\midrule
\multicolumn{2}{l}{\textbf{Intelligence Scale}} \\
\midrule
6 & The robot behaved intelligently while accompanying me. \\
7 & The robot anticipated my movements effectively. \\
8 & The robot handled situations involving multiple people effectively. \\
\bottomrule
\end{tabular}
\end{table}
\subsection{User Study}
\begin{figure*}[t]
\centering
\includegraphics[scale=0.205,page=1]{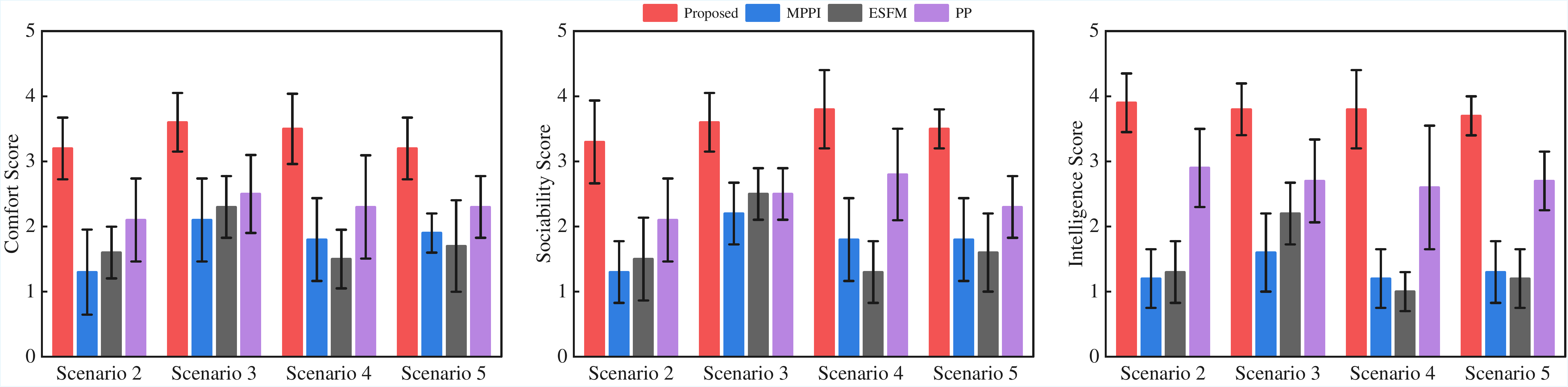}
\caption{User Study Average Scores.}
\label{fig:user_score}
\end{figure*}

A user study was conducted with 20 participants to evaluate their experience while accompanying the robot. Four scenarios were carried out (excluding Scenario 1), and after each trial, participants completed a questionnaire assessing their experience, as summarized in Table~\ref{tab:survey}. The three methods were presented in a randomized order, and participants rated each scenario using a five-point Likert scale. The average scores from the user study are shown in Fig.~\ref{fig:user_score}. The survey results indicate that the proposed method delivered the best overall user experience across all scenarios, consistently receiving the highest scores on the Intelligence metric. Furthermore, the Comfort and Sociability metrics show that users perceived the robot’s behavior under the proposed method as friendlier and safer compared with MPPI, ESFM, and PP.

To verify these observations, a two-way repeated-measures ANOVA was conducted. The results show that the choice of method had a significant effect on all evaluation metrics ($p < 0.05$). Post-hoc analyses further reveal that the Proposed method achieved significantly higher average scores across all three metrics: Comfort, Social, and Intelligence. The scenario factor also exhibited a statistically significant influence on users’ perceived comfort ($p < 0.05$) and sociability ($p < 0.05$). Notably, we observed an interaction effect between Method and Scenario for the Intelligence and Social metrics ($p < 0.05$). This indicates that as the interaction conditions became more complex, the performance gap between the Proposed method and the other methods became more pronounced. These findings confirm that the Proposed method enables the robot to behave more intelligently and naturally, thereby substantially improving the overall user experience across all tested scenarios.
\subsection{Ablation Study}
An ablation study was conducted to evaluate the importance of each component across five scenarios, with results summarized in Table~\ref{tab:ablation}. First, we tested the method without (w/o) the group detection module by VLM to assess its impact. The success rate did not differ significantly, except in Scenario 3, when a member left; the robot continued to treat the departing member as part of the group, which reduced its performance. In Scenario 4, despite the arrival of a new member, the VLM-based group-accompaniment module inferred the new member as an obstacle and adjusted the robot’s position to continue tracking the group while avoiding them, so the performance was not affected.
\begin{table*}[t]
\centering
\caption{Quantitative Results of the Ablation Study}
\label{tab:ablation}
\setlength{\tabcolsep}{3pt}
\resizebox{\textwidth}{!}{
\begin{tabular}{l *{5}{c c c}}
\toprule
\multirow{2}{*}{\textbf{Method}} 
& \multicolumn{3}{c}{\textbf{Scenario 1}}
& \multicolumn{3}{c}{\textbf{Scenario 2}}
& \multicolumn{3}{c}{\textbf{Scenario 3}}
& \multicolumn{3}{c}{\textbf{Scenario 4}}
& \multicolumn{3}{c}{\textbf{Scenario 5}} \\
\cmidrule(lr){2-4}
\cmidrule(lr){5-7}
\cmidrule(lr){8-10}
\cmidrule(lr){11-13}
\cmidrule(lr){14-16}
& SR & CR$\downarrow$ & CD$\downarrow$
& SR$\uparrow$ & CR$\downarrow$ & CD$\downarrow$
& SR$\uparrow$ & CR$\downarrow$ & CD$\downarrow$
& SR$\uparrow$ & CR$\downarrow$ & CD$\downarrow$
& SR$\uparrow$ & CR$\downarrow$ & CD$\downarrow$ \\
\midrule
\rowcolor{gray!20}
\textbf{Proposed}
& \textbf{100} & \textbf{0} & \boldmath{$1.1 \pm 0.04$}
& \textbf{80}  & \textbf{10} & \boldmath{$1.05 \pm 0.06$}
& \textbf{90}  & \textbf{0}  & \boldmath{$1.07 \pm 0.09$}
& \textbf{85}  & \textbf{0}  & \boldmath{$1.03 \pm 0.07$}
& \textbf{95}  & \textbf{5}  & \boldmath{$0.87 \pm 0.11$} \\
w/o VLM group detection
& 100 & 0  & $1.07 \pm 0.07$
& 80  & 0  & $1.10 \pm 0.09$
& 30  & 20 & $2.10 \pm 0.14$
& 85  & 10 & $1.11 \pm 0.09$
& 90  & 10 & $1.06 \pm 0.06$ \\
w/o VLM for companion
& 100 & 0  & $1.03 \pm 0.09$
& 20  & 80 & $1.07 \pm 0.06$
& 0   & 0  & $2.51 \pm 0.17$
& 60  & 40 & $1.06 \pm 0.06$
& 0   & 0  & $1.20 \pm 0.09$ \\
w/o interactive space
& 10  & 60  & $2.09 \pm 0.13$
& 0   & 100 & $2.20 \pm 0.19$
& 0   & 100 & $1.76 \pm 0.13$
& 0   & 100 & $3.50 \pm 0.34$
& 0   & 100 & $1.48 \pm 0.11$ \\
\bottomrule
\end{tabular}
}
\end{table*}
Subsequently, the VLM-based group-accompaniment module was removed, fixing the robot’s accompanying position within the interaction space. In this setup, the robot attempted to maintain a static position and avoid obstacles using only the tracking controller, which led to collisions in all but the first scenario. This highlights the importance of using the VLM for decision-making in robot group accompaniment. Finally, the effect of removing the interaction space input to the VLM was evaluated. It was observed that the interaction space played a critical role in the robot’s performance during accompaniment. Without this input, the robot failed in all scenarios, including the baseline Scenario 1, because the VLM generated inaccurate accompanying positions due to the lack of cues for determining appropriate positions, resulting in collisions and loss of group tracking.
\subsection{Evaluation of Different Group Sizes}
\begin{figure}[t]
\centerline{\includegraphics[scale=0.4,page=1]{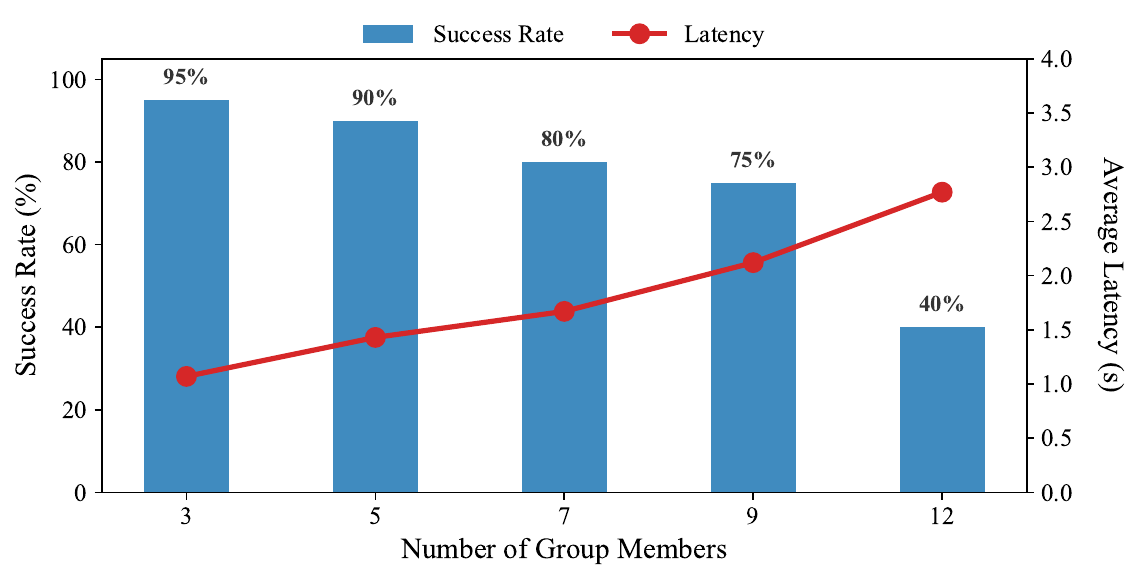}}
\caption{Performance of the proposed method for different group sizes.}
\label{fig:group_size}
\end{figure}
To assess the real-world deployability and scalability of the proposed method, we conducted experiments across multiple group sizes, specifically 3, 5, 7, 9, and 12 members. Each configuration was evaluated over 10 experimental trials. A trial was considered successful if the robot consistently maintained the permitted companionship distance from the group while avoiding any physical collisions. The primary objective of this evaluation was to examine whether increasing the complexity of the human group affects the VLM’s reasoning latency and the robot’s ability to maintain stable behavior when the VLM is integrated with the MPPI-CBF controller.

The experimental results are illustrated in Fig.~\ref{fig:group_size}. The robot demonstrates high success rates in smaller group configurations, achieving 95\% with 3 members and 90\% with 5 members. However, as the group size increases, performance gradually degrades: the success rate decreases to 80\% with 7 members, 75\% with 9 members, and drops significantly to 40\% with 12 members. Furthermore, we observe a clear positive correlation between group size and system latency. The average reasoning latency increases from approximately 1.07 s for 3 members to 1.67 s for 7 members. When the group size reaches 12 members, the latency further increases to approximately 2.77 s. These findings suggest that the framework is most effective for groups of up to 7 members, while larger groups and more crowded environments present challenges that merit further investigation.
\section{Conclusion}\label{sec:conclusion}
In this study, we presented a novel approach enabling a robot to accompany a group of people. By leveraging human-like reasoning and contextual understanding, the robot can detect changes in group membership and adapt to changes in the group’s formation. We introduced an interaction space representation combined with CoT prompting for the VLM, allowing the robot to determine an appropriate accompanying position, which is tracked using an MPPI controller with a CBF. The method was evaluated across five scenarios and compared with previous approaches, alongside a user study. Results show improved tracking performance in terms of success rate and reduced collisions, as well as higher user satisfaction scores. An ablation study confirmed the importance of each component. Future work will involve testing with larger groups, in crowded environments, and using a local VLM to improve reasoning speed.
\bibliographystyle{IEEEtran}
\bibliography{reference.bib}
\end{document}